\documentclass[journal,twoside,web]{ieeecolor}
\usepackage{jsen}
\usepackage{cite}
\usepackage{amsmath,amssymb,amsfonts}
\usepackage{algorithmic}
\usepackage{graphicx}
\usepackage{textcomp}
\usepackage{wrapfig}
\def\BibTeX{{\rm B\kern-.05em{\sc i\kern-.025em b}\kern-.08em
    T\kern-.1667em\lower.7ex\hbox{E}\kern-.125emX}}
\definecolor{abstractbg}{rgb}{0.89804,0.94510,0.83137}
\usepackage{bm}
\usepackage{hyperref}
\usepackage{subfig}
\newcommand{\algname}{\textsc{TrASenD}}

\begin{document}

\title{Attention-Based Deep Learning Framework for Human Activity Recognition with User Adaptation}
\author{Davide Buffelli, Fabio Vandin
\thanks{Work partially supported by MIUR, the Italian Ministry of Education, University and Research, under PRIN Project n. 20174LF3T8 AHeAD (Efficient Algorithms for HArnessing Networked Data) and under the initiative ``Departments of Excellence" (Law 232/2016), and by the grant STARS2017 from the University of Padova.}
\thanks{Davide Buffelli is with the Department of Information Engineering, University of Padova, Padova, Italy (e-mail: davide.buffelli@unipd.it).}
\thanks{Fabio Vandin is with the Department of Information Engineering, University of Padova, Padova, Italy (e-mail: fabio.vandin@unipd.it).}
}

\IEEEtitleabstractindextext{%
\fcolorbox{abstractbg}{abstractbg}{%
\begin{minipage}{\textwidth}%
\begin{wrapfigure}[13]{r}{3in}%
\includegraphics[width=2.8in]{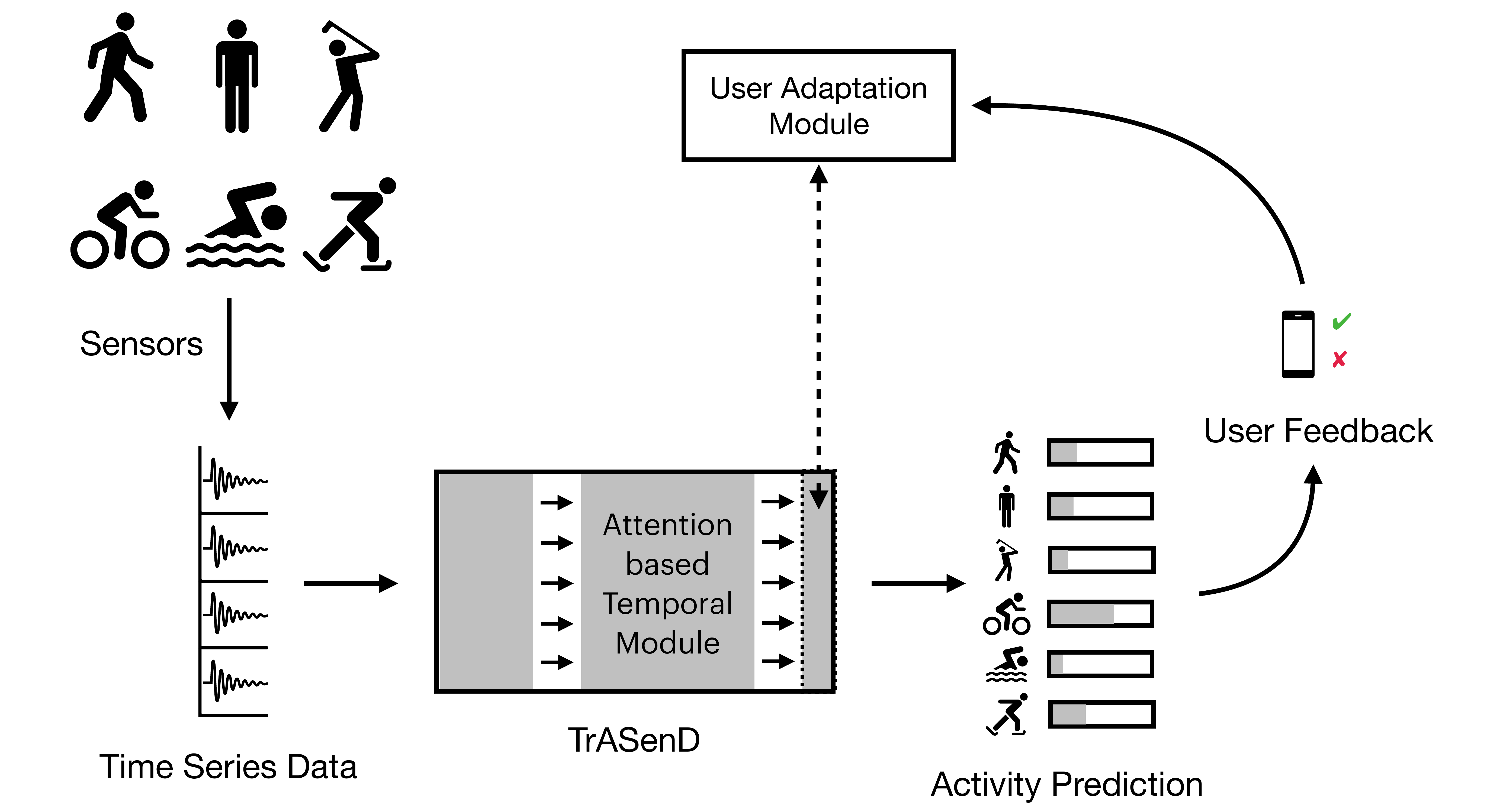}%
\end{wrapfigure}%
\begin{abstract}
Sensor-based human activity recognition (HAR) requires to predict the action of a person based on sensor-generated time series data. HAR has attracted major interest in the past few years, thanks to the large number of applications enabled by modern ubiquitous computing devices. While several techniques based on hand-crafted feature engineering have been proposed, the current state-of-the-art is represented by deep learning architectures that automatically obtain high level representations and that use recurrent neural networks (RNNs) to extract temporal dependencies in the input. RNNs have several limitations, in particular in dealing with long-term dependencies. We propose a novel deep learning framework, \algname, based on a purely attention-based mechanism, that overcomes the limitations of the state-of-the-art. We show that our proposed attention-based architecture is considerably more powerful than previous approaches, with an average increment, of more than $7\%$ on the F1 score over the previous best performing model. Furthermore, we consider the problem of personalizing HAR deep learning models, which is of great importance in several applications. We propose a simple and effective transfer-learning based strategy to adapt a model to a specific user, providing an average increment of $6\%$ on the F1 score on the predictions for that user. Our extensive experimental evaluation proves the significantly superior capabilities of our proposed framework over the current state-of-the-art and the effectiveness of our user adaptation technique. 
\end{abstract}

\end{minipage}}}

© 2021 IEEE. Personal use of this material is permitted. Permission from IEEE must be obtained for all other uses, in any current or future media, including reprinting/republishing this material for advertising or promotional purposes, creating new collective works, for resale or redistribution to servers or lists, or reuse of any copyrighted component of this work in other works.

DOI: 10.1109/JSEN.2021.3067690

\newpage

\maketitle

\section{Introduction}
Sensor-based human activity recognition (HAR) is a time series classification task that involves predicting the movement or action of a person (e.g. walking, running, etc.) based on sensor data. HAR has many practical applications, such as fitness tracking, video surveillance, and gesture recognition. Despite being a well studied and mature problem, HAR has been a very active research area in recent years, due to the rise of ubiquitous computing enabled by smartphones, wearables, and Internet-of-Things devices \cite{NWEKE2018233,94ca4272fbb244418572d2d29c6ef6ae,Bianchi_2019,8950427}.

Several previously proposed approaches tackled the problem by hand-crafting features \cite{Figo2010,Stisen_2015}. These kind of approaches, based on trial-and-error, require a lot of human effort, and therefore time, and are not guaranteed to generalize well to unseen subjects. Deep learning enables automatic feature extraction and can hierarchically compose features to obtain high level representations, which have more discriminative power than handcrafted features based on human expertise. These properties allow deep learning models to be more robust and with higher generalization properties, and have shown great results in HAR \cite{NWEKE2018233,Wang_2019}. In particular, the state-of-the-art is given by the DeepSense framework~ \cite{Yao_2017}, with an architecture based on convolutional neural networks (CNNs) and recurrent neural networks (RNNs).

RNNs have been used in several domains to capture sequential relationships, but present some shortcomings when learning from long input sequences \cite{Hochreiter:01book,Cho_2014}. An attractive strategy to enhance, or replace, RNNs is provided by \emph{attention models}. The main idea behind attention mechanisms is to act as a memory-access mechanism that allows the model to selectively access the most important parts of the input sequence based on the current context. Attention models alleviate RNNs difficulties in learning from long input sequences, and successive developments have led to NLP models based solely on attention mechanisms \cite{Vaswani2017AttentionIA}. To the best of our knowledge, the use of \textit{pure} attention models in deep learning architectures to extract temporal dependencies in multimodal data, such as multi-sensor HAR data, has not been explored.

The human activity recognition task is highly ``personal'', in the sense that a single smartphone or smartwatch is usually used by just one person, and the style of walking, running or climbing stairs is peculiar to each individual. It is then desirable to have deep learning techniques that can be adapted to a specific user. However, the exploration of  personalized deep learning models for HAR has been hitherto ignored.

\subsection{Our Contribution}

We expand the deep learning approaches for HAR with a new purely attention-based framework, \algname, that builds upon the state-of-the-art while significantly outperforming it on three different HAR datasets. \algname\ builds on the observation that RNNs do not provide the best way to capture the temporal relationships in the data, and uses a purely attention-based strategy. We also consider other variants of DeepSense, designed by replacing RNNs with more powerful attention enhanced RNNs mechanisms to capture temporal dependencies, and we show that while they do perform better than DeepSense, they are still less performing than our purely attention based \algname. In addition, we propose a personalization framework to adapt the model to a specific user over time, increasing the accuracy of the predictions for the user. To achieve this result we use a lightweight \textit{transfer learning} approach that continues the training of only a small portion of the model with data acquired from the user. We empirically show that this approach significantly improves the performance of the model on  a specific user.

Our contributions can be summarized as follows:
\begin{itemize}
\item We make use of a purely attention-based mechanism to develop a novel deep learning framework, \algname, for multimodal temporal data.
\item We extensively evaluate \algname\ against the current state-of-the-art and some of its variants that we design. We show that \algname\ significantly outperforms other methods on 3 different HAR datasets, with an average increment of more than $7\%$ on the F1 score over the previous best performing model. We also test the impact of data augmentation, showing that it plays an important role on the generalization capabilities of the models.
\item We propose a new transfer learning technique to adapt a model to specific user, in order to exploit the ``personal'' nature of the HAR task. 
\item We empirically prove the effectiveness of our personalization technique, showing that it leads to an average increment of $6\%$ on the F1 score on the predictions for a specific user. We further show that it is effective on every model we analyze, and on each dataset.
\end{itemize}

\section{Sensing for HAR}
Wearable sensors have now become a common tool for both professional and commercial applications \cite{6974987}. In fact, modern smartphones and smartwatches are equipped with sensors that allow the monitoring of physiological parameters, and the prediction and tracking of physical activities. A practical example of HAR is given by the \textit{fall detection} functionality: given the 3D time series data extracted by an accelerometer, detect if the person has fallen and needs assistance. 

In HAR, sensors usually collect multi-dimensional time series data, which presents important challenges:
\begin{itemize}
\item \textbf{Noise}: data coming from sensors is usually noisy.
\item \textbf{Heterogenous sensing rates}: different sensors may have different sensing rates.
\item \textbf{User generalization and adaptation}: every person has a specific style of walking, running, jumping, etc. It is then important to create systems that are capable of generalizing to new users, but at the same time with the possibility of adapting to the specific style of a given person.
\end{itemize}
The approach proposed in this paper addresses these challenges by: (1) using data augmentation to train models that are robust to noise, (2) preprocessing data to eliminate dependencies on sensing rates, and (3) taking advantage of the generalization capabilities of deep learning models, and further proposing an effective user adaptation procedure.

\section{Related Work}
We divide the previous work related to our contributions in three sections: deep learning approaches for HAR (Section \ref{dlhar}), attention mechanisms (Section \ref{attention}), and  transfer learning and personalization for HAR (Section \ref{tl_and_pers}). 
\subsection{Deep Learning for HAR}\label{dlhar}
Following the taxonomy defined in recent surveys~\cite{NWEKE2018233,Wang_2019}, deep learning techniques for sensor-based HAR fall into three main categories. The first category includes architectures composed of RNNs only (e.g., \cite{Guan_2017,Inoue_2017,Li_2020,9056848}). The second category includes architectures based on CNNs only, and can be further divided in two subcategories of models: \textit{Data Driven} and \textit{Model Driven}~\cite{Wang_2019}. Data Driven models (e.g., \cite{Hammerla2016DeepCA,Sathyanarayana2016ImpactOP,Pourbabaee_2018}) use CNNs directly on the raw data coming from the sensors (each dimension of the data is seen as a channel).
Model Driven approaches (e.g., \cite{Li_2016,Ravi_2016,Singh_2017,8962241,9026890}) first preprocess the data to get a grid-like structure, and then use CNNs. Recent work in the latter category focuses on hybrid models: \cite{9064490} combines multiple CNN models with a fusion layer, that merges the features extracted by the different models, while \cite{9087892} uses a CNN to extract information from sensors, which is then combined with an image segmentation model to produce spinal cord injury predictions.
The third category is represented by those models that use both CNNs and RNNs \cite{Ord_ez_2016,Singh_2017,Yao_2017,Yao_2019,Ma_2019,8918509}.
Finally, other deep learning techniques used for HAR are autoencoders \cite{Wang_2016,almaslukh2017effective}, 
and Restricted Boltzmann Machines \cite{Hammerla2015PDDS,Li_2016_2,Radu_2016}.

DeepSense~\cite{Yao_2017} is a deep learning framework for HAR that belongs to the third category, and constitutes the state-of-the-art for HAR. DeepSense is composed of CNNs to extract features from intervals of data obtained from different sensors, and RNNs (Gated Recurrent Unit (GRU) in particular) to learn temporal dependencies between different time intervals. A final layer is then easily customizable to adapt the framework for classification, regression or segmentation tasks. 

The authors of DeepSense recently proposed a new version of the framework, SADeepSense~\cite{Yao_2019}, where they introduce a \textit{self-attention} mechanism that automatically balances the contributions of multiple sensor inputs. SADeepSense maintains the same architecture of the original DeepSense framework, and adds an attention module to balance the contribution of different sensors based on their sensing quality. Additionally, in the RNN layer, another attention module is used to selectively attend to the most meaningful timesteps. This approach differs significantly from ours as the self-attention module of SADeepSense is used to address the issue of heterogeneity in the sensing quality from multiple sensors, and to select the most relevant timesteps for the final prediction, while \algname\ employs a purely attention-based mechanism directly as a mean to extract temporal dependencies in the data. Furthermore, SADeepSense retains the stacked GRU layer of the original DeepSense framework, while our approach replaces the GRU layer entirely. Another recently proposed architecture based on the DeepSense framework, which adopts a similar attention strategy to SADeepSense is AttnSense \cite{Ma_2019}.

\subsection{Attention Models}\label{attention}
Attention models were first introduced in encoder-decoder neural networks in the context of NLP~\cite{Bahdanau2015NeuralMT}. The main idea behind attention mechanisms is to allow the decoder to selectively access the most important parts of the input sequence based on the current context. This technique serves as a memory-access mechanism, and overcomes RNNs difficulties in learning from long input sequences. Attention has then been used for image captioning in an architecture that made use of both CNNs and RNNs \cite{pmlr-v37-xuc15}. Since then, attention models have become very popular in the deep learning community as an effective and powerful tool to enhance the capabilities of RNNs (e.g., \cite{luong-etal-2015-effective,Chaudhari2019AnAS,Toshevska_2019}). Furthermore, Vaswani et al. \cite{Vaswani2017AttentionIA} introduced the Transformer architecture, which is the current state-of-the-art for NLP, and completely removes RNNs with an attention-only mechanism to model temporal relationships. 

In HAR, attention models have only been used in addition to a RNN (as described in Section \ref{dlhar}), and not as a mean to directly capture temporal dependencies, which is the approach we propose in \algname.

\subsection{Transfer Learning and Personalization in HAR}\label{tl_and_pers}
Transfer learning is not new to HAR. In particular transfer learning has been leveraged to compensate for the amount of labeled data when training a model for activity recognition in different environments/circumstances \cite{Lopes_2011,Cook_2013}. 

A previous (non-deep learning) transfer learning approach for personalized HAR, was proposed by Saeedi et al. \cite{Saeedi_2018}, and used the Locally Linear Embedding (LLE) algorithm to construct activity manifolds, which are used to assign labels to unlabeled data that can be used to develop a personalized model for the target user. Other different approaches to personalized HAR have been made with \textit{incremental learning} \cite{inproceedings} on some classifiers that however were not based on deep learning, and with \textit{Hidden Unit Contributions} \cite{Matsui_2017}, a small layer inserted in between CNNs and learned from user data. In our approach we use transfer learning to train a small portion of the neural network architecture on data provided by a specific user. We show empirically that this simple and easy to implement technique is in fact capable of adapting the framework to the user.
Some preliminary work in this direction can be found in Rokni et al. \cite{Rokni2018PersonalizedHA}. We greatly expand on it by: providing quantitative results on the improvements given by this personalization process; comparing with state-of-the-art techniques; and applying the personalization procedure to multiple, different, deep learning architectures. We also present an empirical evaluation of the learning capabilities of the proposed transfer learning technique.

\section{Data Preprocessing}
\label{sec:preproc}
In this section we present the preprocessing of the sensor measurements that is performed for \algname\footnote{DeepSense \cite{Yao_2017} applies a similar procedure, however, we report some additional details, like the interpolation of the measurements, and the exact values of the parameters, that were not specified in \cite{Yao_2017}.}.
For each sensor $\mathcal{S}^{(i)}$, $i \in \{ 1,...,k \}$, let matrix $\bm{V}^{(i)}$ describe its measurements, and vector $\bm{u}^{(i)}$ define the timestamp of each measurement. $\bm{V}^{(i)}$ has size $d^{(i)} \times n^{(i)}$, where $d^{(i)}$ is the number of dimensions for each measurement from sensor $S^{(i)}$ (e.g., $3$ for both accelerometer and gyroscope as they measure data along the $x$, $y$, and $z$ axes) and $n^{(i)}$ is the number of measurements. $\bm{u}^{(i)}$ has size $n^{(i)}$. For each sensor $\mathcal{S}^{(i)}$, $i \in \{ 1,...,k \}$, the preprocessing procedure is defined as follows:
\begin{itemize}
\item Split the input measurements $\bm{V}^{(i)}$ and $\bm{u}^{(i)}$ along time to generate a series of \emph{non-overlapping} intervals with width $\tau$. These intervals define the set $\mathbb{W}^{(i)} = \{ (\bm{V}^{(i)}_{t}, \bm{u}^{(i)}_{t}) \}$, where $|\mathbb{W}^{(i)}| = T$ and $t \in {1,..., T}$.
\item For each pair belonging to $\mathbb{W}^{(i)}$ apply the Fourier transform and stack the inputs into a $d^{(i)} \times 2f \times T$ tensor $\mathbf{X}^{(i)}$, where $f$ is the dimension of the frequency domain containing $f$ magnitude and phase pairs.
\end{itemize}
Finally, we group all the tensors in the set $\mathbb{X} = \{ \mathbf{X}^{(i)} \}, i \in 1,...,k$, which is then the input to our \algname\ framework.

In practice, we first divide the measurements into samples with a length of 5 seconds \textit{(with no overlap)}, and then apply the procedure with $\tau = 0.25$ seconds and $f = 10$. From now on, with the term \textit{timestep} we refer to a given $\tau$-length interval. In order to deal with uneven sampling intervals that might appear in the data we first interpolate the measurements in each $\tau$-length interval, sample $f$ evenly separated points, and then apply the Fourier transform to those points. The interpolation is done with a linear interpolation along each measurement axis.
The measurements in a 5 seconds sample \textit{of each sensor} are passed to the architecture as a matrix of size \textit{$T \times$ features dimension}, where $T = 20$ and \textit{features dimension} $ = d^{(i)} \times 2f$ (each training and evaluation example is fed to the network with one matrix per sensor). Notice that applying a convolution operation with filters having a receptive field that spans a single row is like extracting features from each $\tau$-length interval separately.

\paragraph*{Data Augmentation}
Similarly to Yao et al. \cite{Yao_2017}, for each training example we added other 9 artificial examples obtained by adding noise (with a normal distribution with zero mean and variance of $0.5$ for the accelerometer and of $0.2$ for the gyroscope). The idea behind this procedure is that the data generated by the sensors are already noisy, so having more samples with slightly different noise should make the network more robust to it. We analyze the impact of data augmentation in our experimental section.

\section{Architecture}
\label{sec:architecture}
In this section we present our framework \algname. We start with a description of the architectural template defined by the DeepSense framework \cite{Yao_2017}\footnote{In \cite{Yao_2017} the authors do not specify several architectural parameters (filter dimensions, strides, presence of padding, dropout probability, training optimizer, learning rate, etc.). We refer to the parameters that can be found on the author's implementation available at \url{https://github.com/yscacaca/DeepSense}.}. We then present the unique characteristics of \algname\ and its redesigned temporal extraction strategy that is based purely on attention. Finally, we present two additional variants of \algname\ with the goal of studying different temporal extraction strategies not based purely on attention, but still more advanced than the stacked GRU layer of DeepSense.

\subsection{DeepSense}\label{ds_arc}

\begin{figure}[h!]
  \centering
  \includegraphics[width=\linewidth]{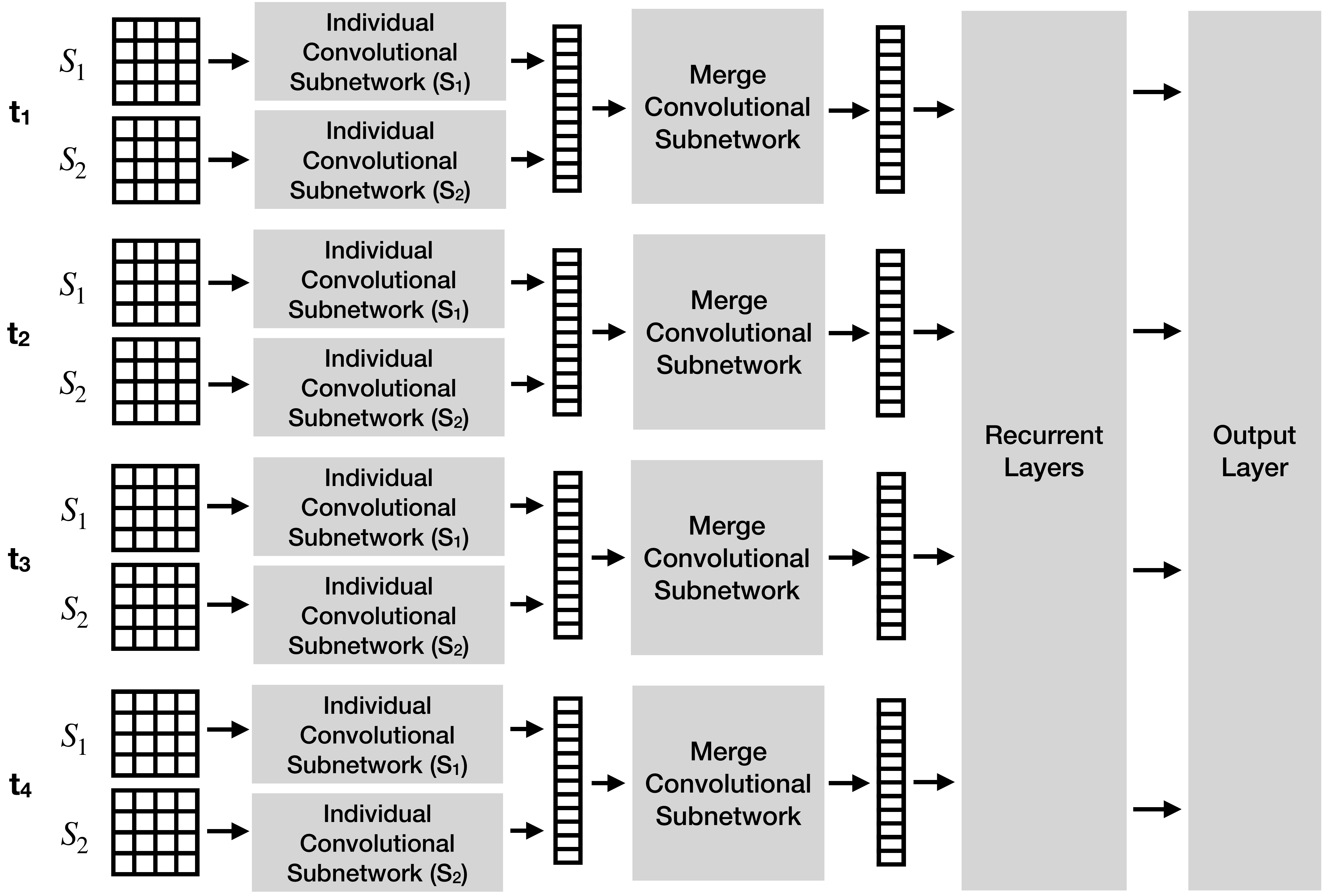}
  \caption{Scheme of the DeepSense framework \cite{Yao_2017}. Individual convolutional subnetworks and the merge convolutional subnetwork share weights across timesteps.}
  \label{fig:ds}
\end{figure}

DeepSense's architecture (Fig. \ref{fig:ds}) can be divided in three parts: \textit{convolutional layers}, \textit{recurrent layers}, and \textit{output layer}. 
The \textit{convolutional layers} can be further divided into two subnetworks: an individual convolutional subnetwork for each sensor and a unique merge convolutional subnetwork. Each individual convolutional subnetwork (one per sensor) takes as input a matrix with dimension \textit{$T \times $ features dimension} (see Section~\ref{sec:preproc}) and is composed of three convolutional layers with 64 filters each. The first layer has filters with dimension $1 \times 6d^{(i)}$ with a stride of  $(1, d^{(i)} \times 2)$ \footnote{Intuitively, the filters have a receptive field that covers three measurement points, and have a stride of one measurement point (after the Fourier transform each point is represented by two numbers: magnitude and phase).}. The second and the third individual convolutional layers have filters with dimension $1 \times 3$. The convolutions in all three layers are applied without padding and are followed by batch normalization \cite{Ioffe:2015:BNA:3045118.3045167}, and a ReLu activation. Furthermore dropout \cite{Srivastava2014DropoutAS} is applied in between the layers, with probability $0.2$. The output of the individual layers are then concatenated, obtaining a tensor with dimension $T \times \text{\textit{number of sensors}} \times features \times channels$ (where \textit{features} depends of the dimension of filters at the previous layers and \textit{channels} is equal to the number of filters of the last individual convolutional layers), and passed to the merge convolutional subnetwork. This subnetwork is composed of three convolutional layers with 64 filters each. For each layer the dimensions of the filters are respectively $1 \times \text{\textit{number of sensors}} \times 8$, $1 \times \text{\textit{number of sensors}} \times 6$, $1 \times \text{\textit{number of sensors}} \times 4$, this time with padding. Again, after each layer, batch normalization and a ReLu activation are performed, with dropout in between layers (with probability $0.2$).

The \textit{recurrent layers} are composed of two stacked GRU \cite{Chung2014EmpiricalEO} layers with 120 cells each. Dropout (with probability $0.5$) and recurrent batch normalization \cite{Cooijmans2017RecurrentBN} are performed between the two layers. Then the mean of the outputs at each time step is taken, and passed to the output layer.

Finally, the \textit{output layer} is a simple dense layer with a number of units equal to the number of activities to predict. The \textit{softmax} activation is used to get a probability distribution between the activities, and cross-entropy is used as loss function: 
\[ L = \sum_{i}^{N} \sum_{c}^{C} -\bm{y}^{(true)}_{i, c} \log(\bm{y}^{(pred)}_{i, c}) \]
where $N$ is the number of training examples, $C$ is the number of different classes, $\bm{y}^{(true)}_{i, c}$ is the $c$-th element of the one-hot encoded ground truth for the $i$-th training example, and $\bm{y}^{(pred)}_{i, c}$ is the $c$-th element of the output of the architecture (after \textit{softmax}) for the $i$-th training example.

\subsection{\algname}
Recurrent Neural Networks (RNNs) present several problems, from the difficulty to learn long-term dependencies \cite{Hochreiter:01book,Cho_2014}, to their low computational efficiency. We propose a new framework, building on the architectural template defined in Section \ref{ds_arc}, that replaces the \textit{stacked-GRU recurrent layer} with an attention-based technique that better exploits temporal dependencies in the data.

\begin{figure*}[h!]
\centering
  \subfloat[]{\includegraphics[height=2.2in]{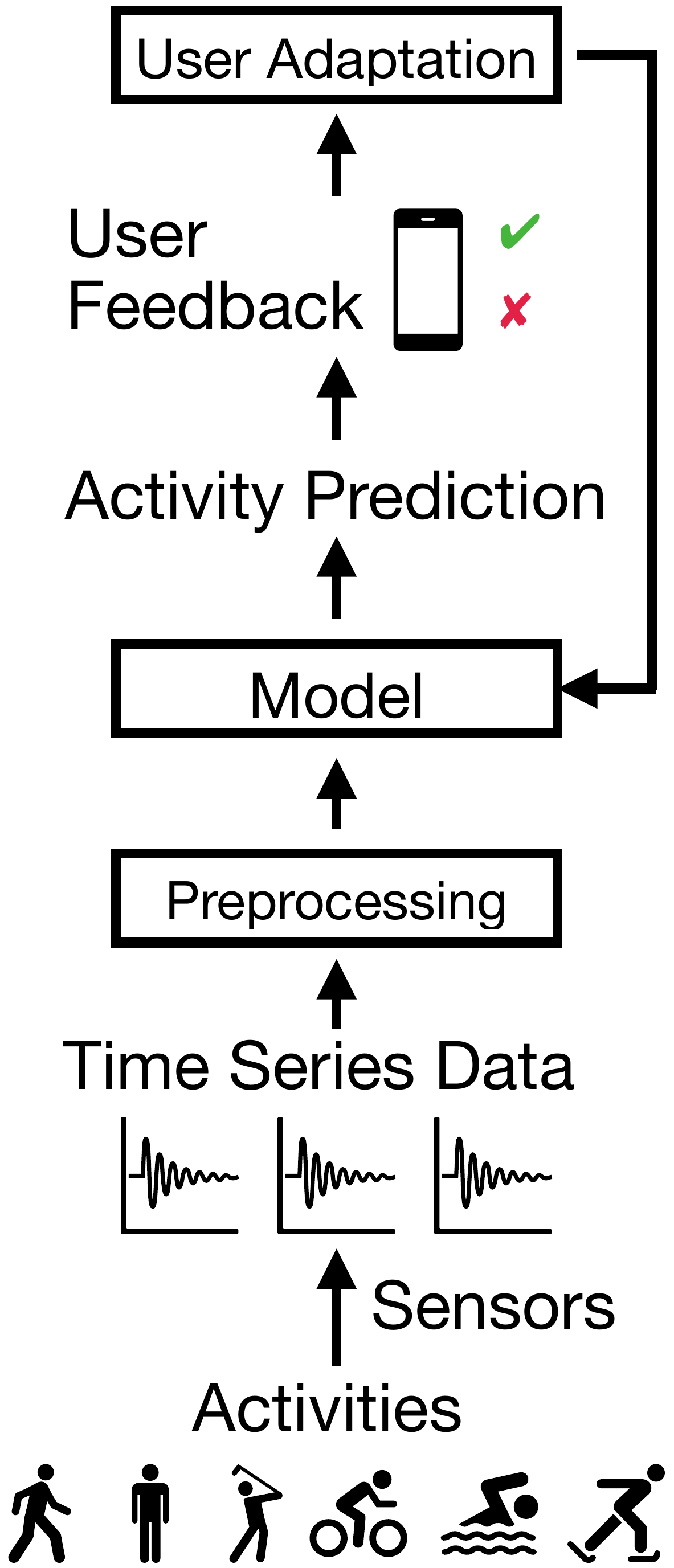}}
  \hspace{\fill} \vline \hspace{\fill}
  \subfloat[]{\includegraphics[height=2.2in]{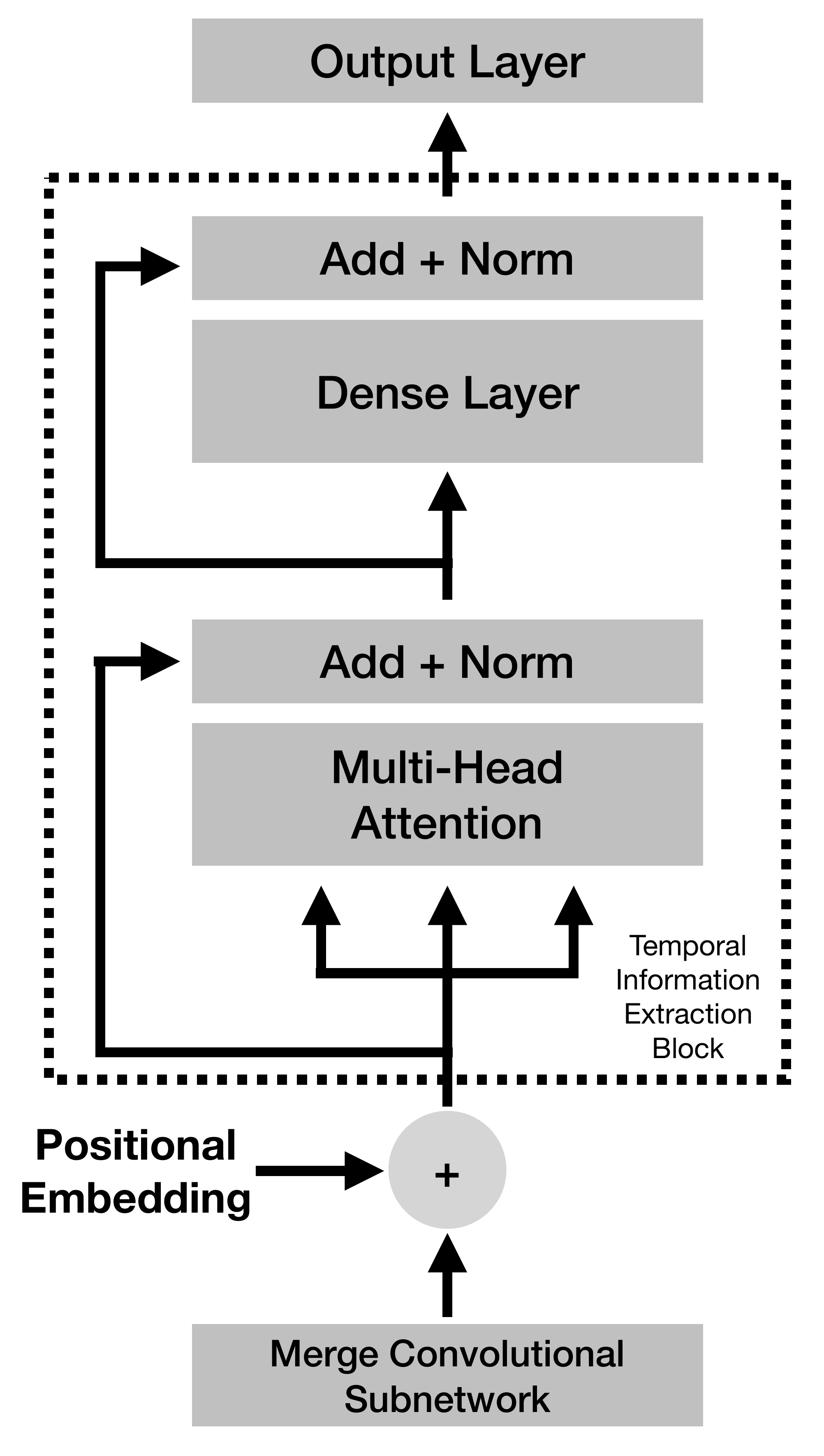}}
  \hspace{\fill}
  \subfloat[]{\includegraphics[height=2.2in]{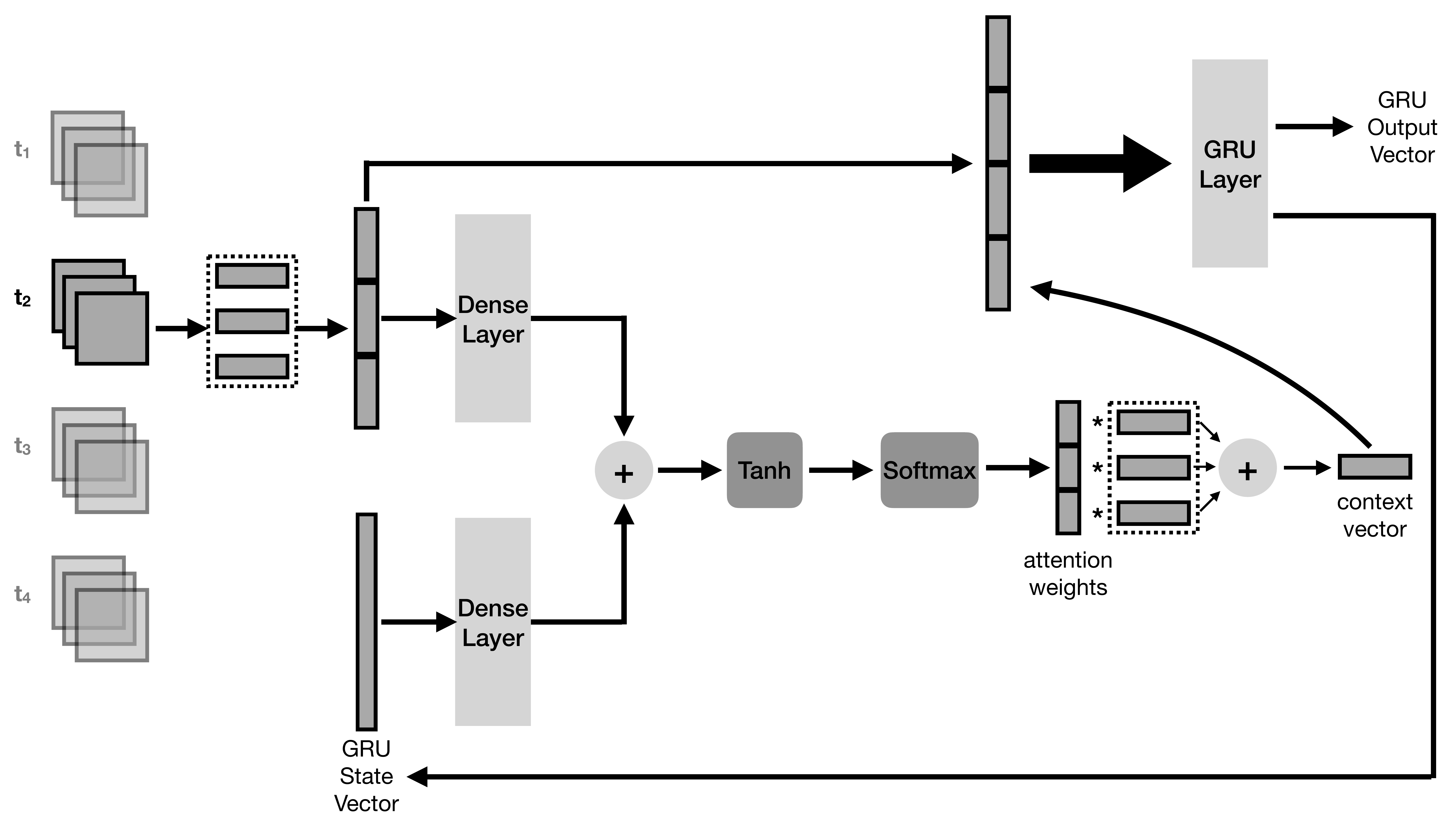}}
  \caption{(a) Flowchart of our method. (b) Scheme of \algname's temporal information extraction block. Notice how temporal information (coming from the Merge Convolutional Subnetwork) is analyzed in a feed-forward manner, without the use of any RNN. 
  (c) Scheme of the attention mechanism for \algname\textit{-CA}. At a given timestep the high level features extracted from the merge convolutional subnetwork are first flattened and concatenated. The attention mechanism, considering the current state of the GRU layer, generates an attention weight for each feature which is then used to scale them. The sum of the scaled features represents the \textit{context vector} which is concatenated to the original features and passed as input to the GRU.}
  \label{fig:ds_variants}
\end{figure*}

We first introduce the attention operator, which is at the core of our attention-based technique for the extraction of temporal dependencies, and then present in more detail the architecture of our proposed framework. Fig. \ref{fig:ds_variants} (b) shows a scheme of the architecture of our temporal dependencies extractor.

\subsubsection{Attention Operator}
An attention operator takes as input three matrices: a \textit{Query} matrix $\bm{Q}$, a \textit{Key} matrix $\bm{K}$, and a \textit{Value} matrix $\bm{V}$, where each row of the matrices indicates the query, key, or value vector of a specific item (where item usually refers to a feature vector). The attention operator attends every query to every key and obtains a similarity score (also called attention score) which is used to obtain weights for all the value vectors (rows of the \textit{Value} matrix). 
Following \cite{Vaswani2017AttentionIA}, we obtain the similarity score using the \textit{scaled dot-product}, and then the attention weights by applying \textit{softmax}. Finally, the values are scaled with their respective attention weight. The whole process can be written as:
\begin{equation*}
\text{\textit{attention}}(\bm{Q}, \bm{K}, \bm{V}) = \text{\textit{softmax}}\left( \frac{\bm{Q}\bm{K}^{T}}{\sqrt{d^{k}}} \right) \bm{V}
\end{equation*}
where $d^{k}$ is the dimension of query and key vectors.
The weights are such that, for every query, the values related to the keys with the highest similarity score are given a higher weight (i.e., more importance). In other words, the weights are used to give more \textit{attention} to the values that are more pertinent to the given query. We talk about \textit{self-attention} when \textit{Query}, \textit{Key}, and \textit{Value} matrices are all referring to items of the same sequence. A \textit{multi-headed} mechanism is such that, for each item, different multiple \textit{Query}, \textit{Key}, and \textit{Value} matrices are created and the attention operator is applied to all of them. The outputs of all the heads are then combined together.

\subsubsection{Architecture}
\algname\ follows the feature extraction procedure and the feed-forward output layer of DeepSense, but completely replaces the \textit{recurrent layers}.
In fact, we only use \textit{attention} to extract temporal dependencies in the data, with a temporal information extractor layer inspired by the Transformer \cite{Vaswani2017AttentionIA}. In more detail, we create a temporal information extractor using a \textit{8-headed self-attention} mechanism. To pass the data to the temporal layer, we reshape the output of the merge convolutional subnetwork to have dimension $T \times \text{\textit{features}}$ (where \textit{features} depends from the size and the number of filters in the merge convolutional subnetwork). The features at different timesteps will be the input of the \textit{self-attention} mechanism. Every sublayer of the temporal block has output with size $T \times \text{\textit{features}}$ to allow residual connections.

We start by applying the positional embedding described by Vaswani et al. \cite{Vaswani2017AttentionIA} to introduce a notion of relative order between the features extracted at different timesteps. Then, for each \textit{head}, we first multiply the input with 3 different learnable matrices to obtain the \textit{query, key, value} matrices $\bm{Q}, \bm{K}, \bm{V}$ (each row of these matrices represents query, key, and value vectors for each timestep). We then obtain the attention score using the \textit{scaled dot-product}, where we used $d^{k} = 64$ and set the dimension of the values to be the same. The attention weights obtained from each \textit{head} are then concatenated and multiplied by a learnable matrix to return to a matrix with dimension $T \times \text{\textit{features}}$. This matrix is then summed with the original inputs (creating a residual connection), and Layer Normalization \cite{Ba2016LayerN} is applied. The data in each timestep is passed through a \textit{position-wise} dense layer\footnote{The same feedforward network is used for each timestep. It is equivalent to a one-dimensional convolutional layer over timesteps with kernel size 1.} with ReLu activation. Finally another residual connection with Layer Normalization is applied to obtain the output of the temporal information extraction block which is then passed to the feedforward output layer. A scheme of the temporal information extraction block can be found in Fig. \ref{fig:ds_variants} (b).

\subsection{Other Architectural Variants}

We now present two variants of \algname\ where we replace the purely attention based temporal information extraction block, with other (simpler, but more advanced than regular RNNs) techniques to capture temporal dependencies in the input.

\paragraph{\algname-BD} The first variant substitutes the pure attention temporal block with a bidirectional-RNN (BRNN) \cite{Schuster_1997}. A BRNN generalizes the concept of RNNs by connecting two hidden layers of opposite directions to the same output (we continue using GRUs as forward and backward hidden layers). This allows the network to get information from past and future inputs simultaneously. At each timestep we now get the state of both forward and backward cells, so we concatenate them, and finally take the average of the concatenated outputs at each timestep and pass them to the output layer. 
\paragraph{\algname-CA} Inspired by the work by Xu et al. \cite{pmlr-v37-xuc15}, we use a GRU layer (we keep it with 120 cells) with an attention mechanism over the output features of the merge convolutional subnetwork. We first average the features extracted from the first $\tau$-length interval (first \textit{timestep}) and pass it through a dense layer to obtain the initial state for the GRU layer. We then use the following attention mechanism: at each timestep, we pass the features extracted by the CNN layers and the current state of the GRU through two different dense layers \textit{without applying any activation function}. We then sum the two outputs and apply $tanh$ before passing it to \textit{softmax} to obtain the \textit{attention weights}. Finally, the features are scaled with their attention weights. The sum of the scaled feature vectors forms the \textit{context vector} which is then concatenated to the original features for the current timestep and passed as input to the GRU. A scheme of this attention mechanism can be found in Fig. \ref{fig:ds_variants} (c). The rest of the architecture remains unchanged.

\subsection{Transfer Learning Personalization} \label{sec:tl_pers}
To make the system capable of adapting to a specific user over time, we propose a simple \textit{transfer learning} strategy (Figure \ref{fig:ds_variants} (a)). Transfer learning is a method where a model developed for a task is reused as the starting point to learn a model on a second task. The typical scenario in a transfer learning setting is to have a trained base network, which is repurposed by training on a target dataset. The idea is that the pre-trained weights in the base network can ease the training on the target dataset. We slightly depart from this scenario by extracting the output layer from a trained \algname\ model (and other proposed variants); that is, we are using transfer learning only on the output layer. More in detail, the data coming from the sensors will be passed to the \algname\ architecture, up to the end of the temporal layer. The output layer becomes a separate network that receives the output of the temporal layer as input, and will be trained with the data generated by the user. This can be implemented in a practical scenario by first using a model trained on one of the datasets, and after each prediction, asking the user to manually insert the activity he was performing. We then use these new data samples to retrain \textit{only} the output layer, which is a single layer dense network that can easily be trained on-device. This procedure allows the architecture to take advantage of the complex general \textit{feature extracting} mechanism that reduces multimodal time series to a fixed size vector, and to successively learn user-specific feature characteristics.

\section{Experimental Evaluation}
We present here the datasets and the procedure used to evaluate the performance of \algname, and the effectiveness of the proposed personalization process.

\subsection{Datasets}
We present below the three HAR datasets used in our tests. Our choices were based on the statistics shown in Table 3 of the survey by Wang et al. \cite{Wang_2019}: we consider the datasets that have data from at least 9 subjects (to better test generalization properties), with at least 2 different sensing modalities (to test the various methods on multimodal data), and then take the datasets with the largest number of samples. A summary of the chosen datasets can be found in Table \ref{tab:dataStat}.

\paragraph{HHAR \cite{Stisen_2015}} The Heterogeneity Activity Recognition Data Set contains data from accelerometer and gyroscope of 12 different devices (8 smartphones and 4 smartwatches) used by 9 different subjects while performing 6 activities. We only considered data coming from smartphones.
\paragraph{PAMAP2 \cite{Reiss_2012_2,Reiss_2012}} The Physical Activity Monitoring dataset contains data of 12 different physical activities, performed by 9 subjects wearing 3 inertial measurement units and a heart rate monitor. We only considered data coming from the inertial measurement units (IMU), which were positioned in three different body areas (hand, chest, ankle) during the measurements. From each IMU we considered data measured by the first accelerometer, the gyroscope and the magnetometer. This provides a scenario with data coming from 9 input sensors.  
\paragraph{USC-HAD \cite{Zhang_2012}} The University of Southern California Human Activity Dataset uses high precision specialised hardware, and has a focus on the diversity of subjects, balancing the participants based on gender, age, height and weight. The dataset contains measurements from accelerometer and gyroscope obtained from 14 different subjects while performing 12 activities.

\begin{table}[h!]
  \centering
    \caption{Summary of the multi-modal HAR datasets used for our tests.}
    \label{tab:dataStat}
    \begin{tabular}{lccc}
      \hline
      \textbf{Dataset} & \textbf{Subjects} & \textbf{Activities} & \textbf{Input Sensors}\\
      \hline
      \textbf{HHAR} & 9 & 6 & 2\\
      \textbf{PAMAP2} & 9 & 12 & 9\\
      \textbf{USC-HAD} & 14 & 12 & 2\\
      \hline
    \end{tabular}
\end{table}

\subsection{Baselines}
We choose an extensive collection of deep learning, and non-deep learning methods to compare to \algname\ and its variants. For all considered models, we use the implementation provided by the authors when available. Unless otherwise specified we use the model hyperparameters defined by the authors.

\paragraph{Deep Learning Baselines} We test our algorithm against all the Deep\-Sense-based architectures, and additional deep learning techniques. In particular for the Deep\-Sense-based architectures we test against the original DeepSense \cite{Yao_2017}, and the two latest attention enhanced versions: SADeepSense \cite{Yao_2019}, and AttnSense \cite{Ma_2019}.
We then consider DeepConvLSTM \cite{Ord_ez_2016} which is a CNN+LSTM approach, and its new attentive version proposed in \cite{Murahari_2018} that we call DeepConvLSTM-Att. All the attention models considered thus far add an attention module to a RNN layer, while we remember that our algorithm \algname\ completely removes RNNs in favour of a purely attention-based temporal information extraction technique.
We also provide some results for a basic LSTM based architecture (we implement it with 2 LSTM layers, each with 256 cells, followed by a fully connected layer that outputs the predicted class). Finally, to take into consideration also other deep learning techniques we consider MultiRBM \cite{Radu_2016}, where a Restricted Boltzman Machine (RBM) is used for each sensor, and a single final RBM is used to then merge all the outputs for the sensors and obtain the predicted class.

\paragraph{Non-Deep Learning Baselines} As non-deep learning baselines we considered a Random Forest (RF) classifier (one of the most used and most effective shallow classifiers for HAR \cite{Stisen_2015}) on the same raw frequency domain features fed to the deep learning approaches (denoted with  \textbf{RF-FF}), and then on the most used handcrafted frequency domain features (DC Component, Spectral Energy, and Information Entropy; denoted with \textbf{RF-HC}). 

\subsection{Experimental Setup}
For all tests we performed \textit{leave-one-user-out} cross validation: we train on data from all subjects except one, and we use the data from the excluded subject as test set. We perform this procedure for each subject and then average the results. This validation procedure follows the common practices in the field, and ensures that the model is not overfitting to the training data.

To evaluate the personalization process we divide the data of each activity of the excluded user into two equal \textit{time-contiguous} parts. One part is used to personalize the output layer after the model has been learned on all other users, and the other is used as test set. We also make sure to feed the data, both for training and validation, in time-contiguous samples (simulating the real-world personalization procedure described in Section \ref{sec:tl_pers}).

Due to the imbalance in the number of samples per-class we use the F1 score as the measure to quantify the performance of the models. All \algname\ models were implemented\footnote{Code is available at: \url{https://github.com/DavideBuffelli/TrASenD}} using TensorFlow \cite{abadi2016tensorflow}.

To ensure a fair comparison and to avoid ``hyperparameter hacking'' we kept all the values for the \textit{architecture hyperparameters} (filters size, dropout probability, number of filters, number of GRU units, etc.: see Section~\ref{sec:architecture}.) equal for each DeepSense-based model. Furthermore, for all models, the only optimized hyperparameter was the \textit{learning rate}. To do so we took out 1 user and tried the training and evaluation procedure on the HHAR dataset, with \textit{learning rate}  $\in {\{10^{-2}, 10^{-3}, 10^{-4}\}}$. We then considered the setting that gave the highest F1 score on the user's data and used it for \textit{all} datasets (no optimization for each different dataset).
In the training procedure we trained for 30 epochs for each user and took the model of the epoch with the highest performance. All \algname\ based models were trained using the Adam Optimizer~\cite{article}. The other methods were trained with the optimization technique suggested by the authors.
For the personalization process, we retrain the output layer for 1 epoch (per each new data point separately) with TensorFlow's default Adam optimizer parameters: $\alpha = 0.001$, $\beta1 = 0.5$, $\beta2 = 0.9$, and $\epsilon = 10^{-8}$.

\subsection{Results}
\begin{figure}[h!]
  \centering
  \includegraphics[width=0.8\linewidth]{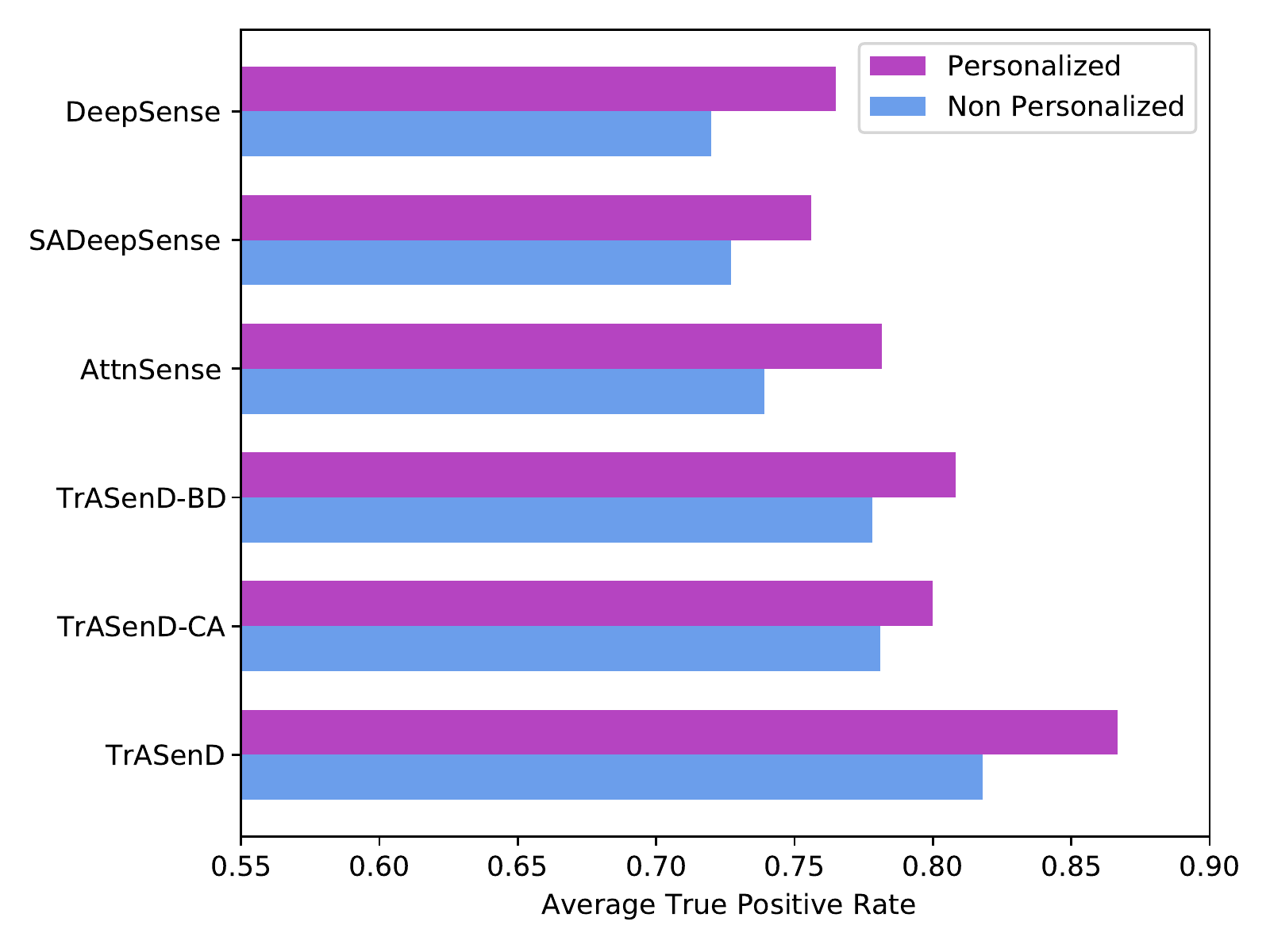}
  \caption{Personalization increases the average True Positive rate of the deep learning models.}
  \label{fig:tp_rate}
\end{figure}

\begin{table}[h!]
  \centering
    \caption{F1 score results on different HAR datasets.}
    \label{tab:res2}
    \begin{tabular}{lcccccc}
      \hline
      \textbf{Model} & \multicolumn{3}{c}{\textbf{Dataset}}\\
      & HHAR & PAMAP2 & USC-HAD\\
      \hline
      RF-FF                                                         & 0.569 & 0.512 & 0.417 \\
      RF-HC \cite{Stisen_2015}                          & 0.575 & 0.501 & 0.474 \\
      MultiRBM \cite{Radu_2016}                       & 0.647 & 0.589 & 0.598 \\
      LSTM                                                          & 0.663 & 0.583 & 0.612 \\
      DeepConvLSTM \cite{Ord_ez_2016}         & 0.701 & 0.633 & 0.658 \\
      DeepConvLSTM-Att \cite{Murahari_2018} & 0.735 & 0.647 & 0.682 \\
      DeepSense \cite{Yao_2017}                       & 0.720 & 0.647 & 0.670 \\
      SADeepSense \cite{Yao_2019}                  & 0.753 & 0.661 & 0.688 \\
      AttnSense \cite{Ma_2019}                          & 0.762 & 0.657 & 0.685 \\ 
      \textbf{\algname\textit{-BD}}                                  & 0.798 & 0.650 & 0.681 \\
      \textbf{\algname\textit{-CA}}                                  & 0.797 & 0.659 & 0.687 \\
      \textbf{\algname}                                        & \textbf{0.848} & \textbf{0.723} & \textbf{0.702} \\
      \hline
    \end{tabular}
\end{table}
Table \ref{tab:res2} summarizes the F1 score results for \algname\ and the other methods we considered, on the three datasets. We can observe that \algname\ and its variants present higher F1 score than DeepSense on all the three datasets. Furthermore we notice that \algname\ always achieves the highest performance with a big margin. In fact, \algname\ shows an F1 score that is, on average, $7\%$ higher then the previous best performing model. These results confirm that our attention-based technique (without RNNs) is highly capable of extracting temporal dependencies. Most notably we can see that \algname\ significantly outperforms the newer SADeepSense and AttnSense, whose performance are comparable to the ones of \algname\textit{-BD} and \algname\textit{-CA}, which are far from \algname's.
In Figure \ref{fig:tp_rate} we show how the average True Positive rate is affected by the personalization process on the HHAR dataset. We notice an average 5\% increase, further confirming the ability to adapt to a specific user.
We remark that all the results come from a cross validation procedure where the test data is coming from a user that was not seen during training, hence showing that the model is not simply overfitting the training data.

\begin{table}[h!]
  \centering
    \caption{F1 score results of the deep learning models with (P) and without (NP) personalization.}
    \label{tab:res3}
    \begin{tabular}{lcccccc}
      \hline
      \textbf{Model} & \multicolumn{6}{c}{\textbf{Dataset}}\\
      & \multicolumn{2}{c}{HHAR} &  \multicolumn{2}{c}{PAMAP2} &  \multicolumn{2}{c}{USC-HAD}\\
      & NP & P & NP & P & NP & P\\
      \hline
      DeepSense \cite{Yao_2017}                       & 0.720 & 0.775 & 0.647 & 0.693 & 0.670 & 0.712\\
      SADeepSense \cite{Yao_2019}                  & 0.753 & 0.790 & 0.661 & 0.699 & 0.688 & 0.749\\
      AttnSense \cite{Ma_2019}                          & 0.762 & 0.801 & 0.657 & 0.689 & 0.685 & 0.746\\ 
      \textbf{\algname\textit{-BD}}                                  & 0.798 & 0.821 & 0.650 & 0.699 & 0.681 & 0.748\\
      \textbf{\algname\textit{-CA}}                                  & 0.797 & 0.819 & 0.659 & 0.701 & 0.687 & 0.726\\
      \textbf{\algname}                                        & \textbf{0.848} & \textbf{0.889} & \textbf{0.723} & \textbf{0.749} & \textbf{0.702} & \textbf{0.759}\\
      \hline
    \end{tabular}
\end{table}

Table \ref{tab:res3} presents the F1 score of the DeepSense-based models when evaluated on the datasets with and without applying personalization. The results confirm the effectiveness of our transfer learning personalization process giving an average $\approx6.2\%$  increase on the F1 score independently of dataset and base architecture.

These results confirm that restricting the transfer learning to the last layer of the network allows the model to retain the generalization capabilities in the extraction of useful feature (hence confirming the robustness to overfitting), while allowing the last layer to adapt to a specific user.

\subsubsection{Validating the Personalization Process} To prove that the training of the output layer alone can significantly impact on the performance of the network we first train the full model of Section \ref{ds_arc} on the HHAR dataset with randomly permuted labels, and then we perform the personalization process on correctly labeled data. The resulting F1 scores (on the test set) are 0.166 and 0.523, respectively.
We can notice that the model trained on data with randomly permuted labels has the performance of a uniform random classifier, as one would expect, and the personalization process is capable of significantly boosting the performance of the model. This result shows that in fact the re-training of the output layer alone can largely affect the outcome of the model. 
\subsubsection{Impact of Data Augmentation}
\begin{figure}[h!]
  \centering
  \includegraphics[width=0.9\linewidth]{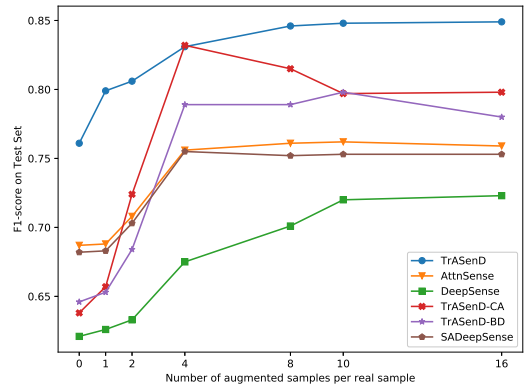}
  \caption{Performance of the deep learning models on HHAR when trained with different number of augmented samples.}
  \label{fig:aug_plot}
\end{figure}

To asses the benefits of the data augmentation procedure, we evaluate all the deep learning models based on the DeepSense framework on HHAR with and without augmented data. The results, shown in Table \ref{tab:resAug}, confirm that data augmentation is important to train a model that is more robust to noise, and in fact we can see a significant increase in the F1 score. Fig. \ref{fig:aug_plot} shows how the performance of the analyzed DeepSense variants change when trained with different number of augmented samples. It's interesting to see that using 4 augmented samples for each real sample, already provides an important performance gain. We also notice that \algname\ is always superior to the other architectures, and performs significantly better than the others even when trained without augmented samples. Furthermore, we see that SADeepSense and \algname\ are the two architectures showing the smallest gap between highest and lowest F1 score result, confirming their superior generalization properties, with \algname\ achieving a higher overall F1 score.
\begin{table}[h!]
  \centering
    \caption{Performance on HHAR with (A) and without (NA) data augmentation.}
    \label{tab:resAug}
    \begin{tabular}{lcc}
      \hline
      \textbf{Model} & \multicolumn{2}{c}{\textbf{F1 Score on Test Set}}\\
      & A & NA\\
      \hline
      DeepSense \cite{Yao_2017}      & 0.720 & 0.621 \\
      SADeepSense \cite{Yao_2019} & 0.753 & 0.682 \\
      AttnSense \cite{Ma_2019}          & 0.762 & 0.687 \\
      \algname\textit{-BD}                     & 0.798 & 0.646\\
      \algname\textit{-CA}                     & 0.797 & 0.638 \\
      \textbf{\algname}                       & \textbf{0.848} & \textbf{0.761} \\
      \hline
    \end{tabular}
\end{table}

\section{Conclusions}
In this paper we presented \algname, a new deep learning framework for multimodal time series, and also proposed a transfer learning procedure to personalize the model to a specific user for the human activity recognition tasks. \algname\ is designed to improve the extraction of temporal dependencies in the data by replacing RNNs with a purely attention based temporal information extraction block. Our extensive experimental evaluation shows that \algname\ significantly outperforms the state-of-the-art and that, in general, replacing RNNs with attention-based strategies leads to significant improvements. In particular, we obtain an average increment of more than $7\%$ on the F1 score over the previous best performing model. We also show the effectiveness of our simple personalization process, which is capable of an average $6\%$ increment on the F1 score on data from a specific user, and the impact of data augmentation.

The personalization procedure we propose may impact the user experience while using an application that implements our technique. In fact, asking too many times for feedback about the model's predictions may not be feasible. Future research directions include the optimization of the personalization process to minimize the feedback required from the user, for example by using data augmentation or curriculum training techniques \cite{Bengio_2009}.

\bibliographystyle{plain}
\bibliography{scibib}

\begin{IEEEbiography}[{\includegraphics[width=1in,height=1.25in,clip,keepaspectratio]{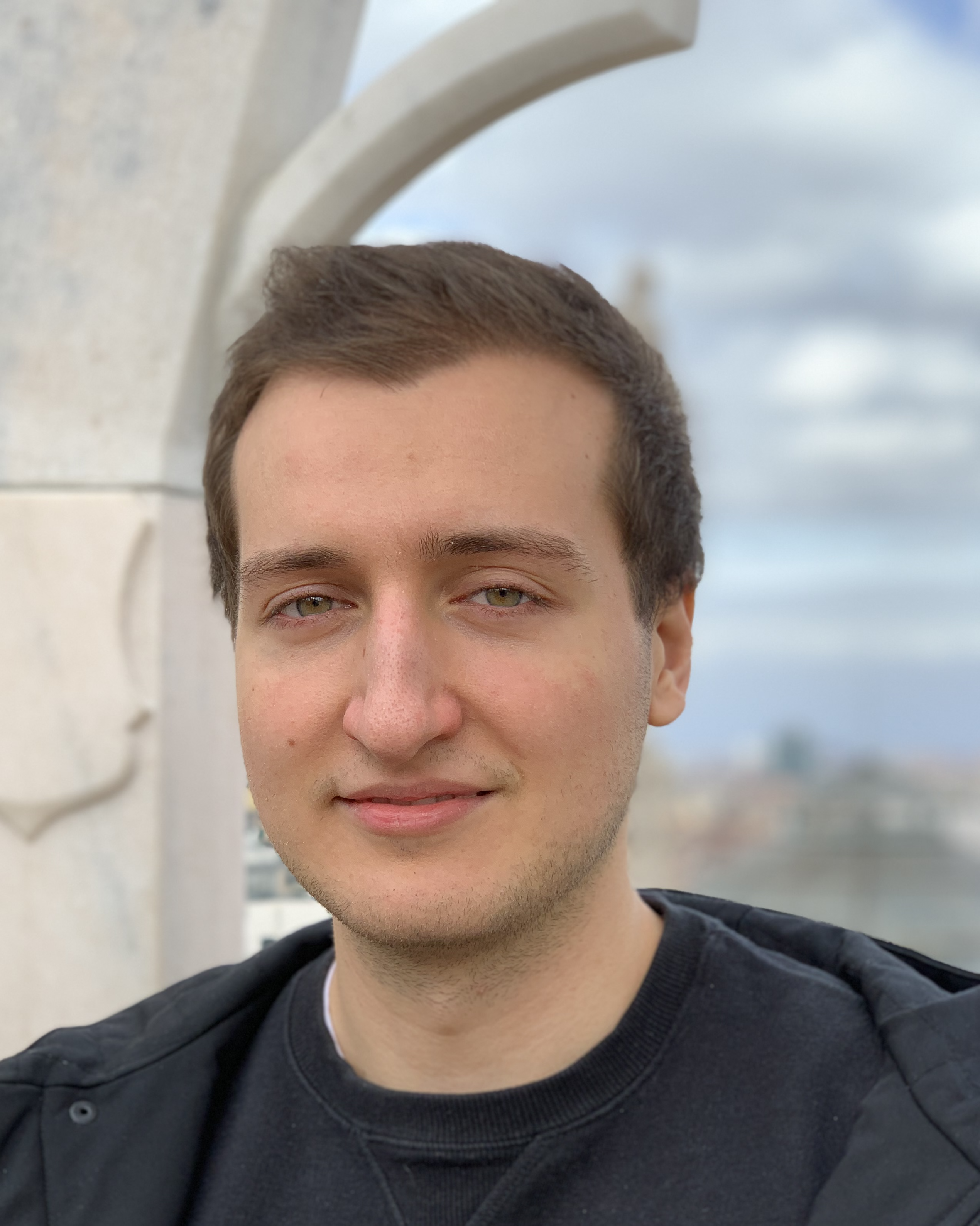}}]{Davide Buffelli} was born in Verona, Italy in 1994. 
He received the B.S. degree in Information Engineering in 2016, and the M.S. degree in Computer Engineering in 2019, both from the University of Padova, Padova, Italy.
He is currently pursuing the Ph.D. degree in Information Engineering at the University of Padova, Padova, Italy.

From June, to December 2018, he was a Data Science Intern at Philips Digital and Computational Pathology. From April 2019 to September 2019 he was a graduate 
Research Fellow at the University of Padova.
His research interests lie in the area of Deep Learning, with a focus on techniques for temporal data, and graph structured data. 
\end{IEEEbiography}

\begin{IEEEbiography}[{\includegraphics[width=1in,height=1.25in,clip,keepaspectratio]{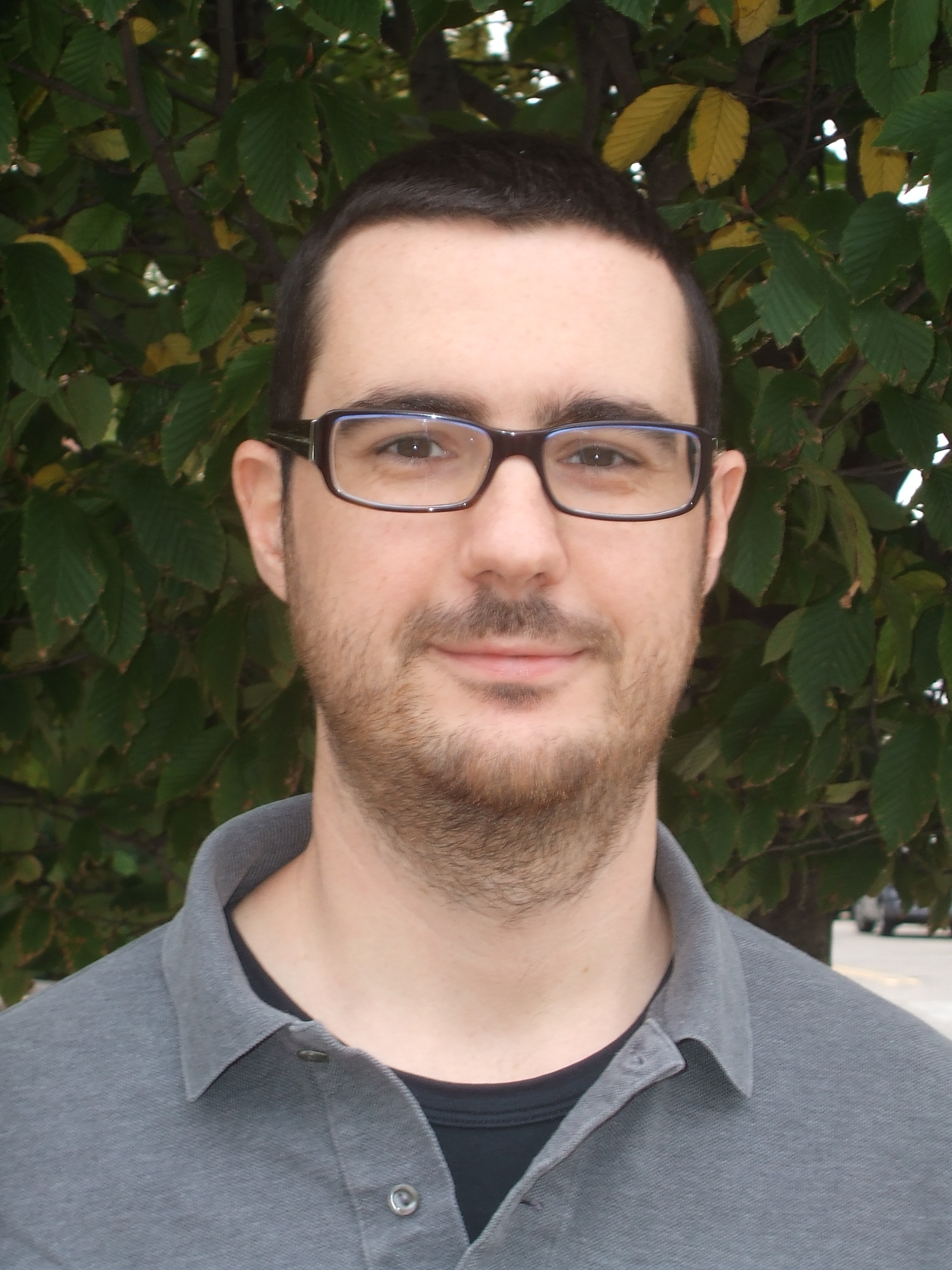}}]{Fabio Vandin} was born in
Soave, Italy, in 1982. He received the B.S. degree in Computer Engineering in 2004, the M.S. degree in Computer Engineering in 2007, and the Ph.D. in Information Engineering in 2010, all from the the University of Padova, Italy. 
Since 2020, he has been a Professor at the Department of Information Engineering at the University of Padova, Italy. His main research interests are in the area of algorithms for data mining and machine learning, and 
applications to biomedicine, molecular biology, and e-health.

He has been an Assistant Professor Research at Brown University, RI, USA, an Assistant Professor at the University of Southern Denmark, Odense, Denmark, and an Associate Professor at the University of Padova, Italy. In 2016 he has been a Research Fellow at the Simons Institute for the Theory of Computing at UC Berkeley, USA.
He has authored more than 60 papers in international peer-reviewed conferences and journals.
\end{IEEEbiography}

\end{document}